\documentclass[10pt,twocolumn,letterpaper]{article}

\usepackage[pagenumbers]{cvpr}      
\usepackage{multirow}
\usepackage{float}
\usepackage{amsmath}
\usepackage{array}
\usepackage{tabularx}
\usepackage{makecell}
\usepackage[dvipsnames]{xcolor}

\definecolor{cvprblue}{rgb}{0.21,0.49,0.74}
\usepackage[pagebackref,breaklinks,colorlinks,allcolors=cvprblue]{hyperref}

\def\paperID{74} 
\def\confName{3DV\xspace}
\def\confYear{2026\xspace}

\title{Tessellation GS: Neural Mesh Gaussians for Robust Monocular Reconstruction of Dynamic Objects}

\author{
  Shuohan Tao\textsuperscript{1}, 
  Boyao Zhou\textsuperscript{2}, 
  Hanzhang Tu\textsuperscript{2}, 
  Yuwang Wang\textsuperscript{2}, 
  and Yebin Liu\textsuperscript{2}\thanks{Corresponding author (liuyebin@mail.tsinghua.edu.cn).} \\ 
  \bigskip 
  \small 
  \textsuperscript{1}University of Cambridge,
  \textsuperscript{2}Tsinghua University
}
\begin{document}
\maketitle
\begin{abstract}
3D Gaussian Splatting (GS) enables highly photorealistic scene reconstruction from posed image sequences but struggles with viewpoint extrapolation due to its anisotropic nature, leading to overfitting and poor generalization, particularly in sparse-view and dynamic scene reconstruction. We propose Tessellation GS, a structured 2D GS approach anchored on mesh faces, to reconstruct dynamic scenes from a single continuously moving or static camera. Our method constrains 2D Gaussians to localized regions and infers their attributes via hierarchical neural features on mesh faces. Gaussian subdivision is guided by an adaptive face subdivision strategy driven by a detail-aware loss function. Additionally, we leverage priors from a reconstruction foundation model to initialize Gaussian deformations, enabling robust reconstruction of general dynamic objects from a single static camera, previously extremely challenging for optimization-based methods. Our method outperforms previous SOTA method, reducing LPIPS by 29.1\% and Chamfer distance by 49.2\% on appearance and mesh reconstruction tasks.
\end{abstract}    
\section{Introduction}
\label{sec:intro}
Reconstruction of observed scenes has always been a major challenge in computer vision. With the spark of differentiable rendering, the task has shifted from solely relying on special equipments like depth cameras and LiDARs to leveraging more accessible multi-camera or monocular video setup. However, most methods~\cite{newcombe2011kinectfusion,mildenhall2020nerf,kerbl2023_3dgs} focus on static scenes due to the inherent under-determinacy of dynamic scene reconstruction and non-rigid deformation.

Early approaches~\cite{chen1993view,newcombe2015dynamicfusion} employed image warping to preserve observed information for novel view synthesis or physical plausibility regularization in unobserved regions for geometry reconstruction. 
Recent monocular differentiable rendering methods incorporate geometric constraints alongside photometric loss with implicit NeRF~\cite{luiten2023dynamic, huang2024sc, yang2024deformable, guo2024motion} or explicit Gaussian Splatting~\cite{lin2024gaussian,Wu2024_4dgs,liu2024dynamic} representation.
Such "4D" methods leverages spectral-biasedness of MLP~\cite{Wu2024_4dgs, yang2024deformable, das2024neural} or spatial-temporal factorization~\cite{fang2022TiNeuVox, cao2023hexplane, park2021nerfies, Wu2024_4dgs} but relies on explicit or implicit (camera movement) multi-view input.
As pointed out by~\cite{gao2022monocular}, existing differentiable rendering approaches are highly susceptible to view overfitting, and using unrealistic camera motions in datasets D-NeRF~\cite{pumarola2021d} and DG-Mesh~\cite{liu2024dynamic}.
On the other hand, avatar-like methods~\cite{hu2024gaussianavatar, guo2023vid2avatar, zheng2023pointavatar, li2024animatable} enable per-pose Gaussian attributes modeling for human performance reconstruction, which heavily rely on 
category-specific template, \textit{e.g.} SMPL~\cite{loper2015smpl}. 
Thus, such methods are not able to handle topological changes and category-agnostic reconstruction.

In terms of template-free method, DG-Mesh~\cite{liu2024dynamic}, TiNeuVox~\cite{fang2022TiNeuVox}, and HexPlane~\cite{cao2023hexplane} pose various structural and loss designs.
DG-Mesh utilized Laplacian regularization to ensure smoothness of the reconstructed geometry. TiNeuVox and HexPlane factorizes spatiotemporal information into several feature planes that are decoded by MLP. Although effective for input video with enough camera movement, they do not explicitly regularize view-overfitting, and as a result, they can not recover dynamic details from general monocular video input and degrade synthesis quality for novel views far from input view, due to the lack global cue of moving object.

To bridge this gap, we propose to use a template-free geometry prior to anchor our surfel-like Gaussian attributes.
To obtain such prior, we apply large reconstruction model (LRM) to extract coarse geometry frame by frame.
Such geometry is in a relatively low resolution without high-frequency details and correspondence between them is unknown.
Therefore, we propose to learn a deformation field with control points whose movements are defined by a motion MLP to build the correspondence between reference frame and other frames and to preserve temporal consistency by doing a sequential optimization.
In this process, we design a robust Chamfer loss to overcome the flickering geometry and floating artifacts from LRM output.
We only optimize deformation field in this stage, while further anchor 2D Gaussian attributes for joint optimization of deformation and appearance in stage 2.
Since the deformation field is established in stage 1, we define all Gaussian points on the geometry surface of reference frame and learn Gaussian attributes with appearance decoders and Gaussian features defined on mesh vertices.
To further enhance the high-frequency appearance modeling, we propose a novel hierarchical structure of Gaussian points, which follows a structural densification mechanism instead of the gradient-based splitting in original Gaussian Splatting. Attaching surfel-like Gaussians on geometry triangles, our appearance decoders are able to optimize both geometry and appearance with solely input images.
Finally, we achieve accurate motion reconstruction, temporally consistent geometry, and photo-realistic appearance modeling within 90 minutes for 500 frames.


In conclusion, our main contributions are as follows: 
\begin{itemize}
    \item Our method is capable of reconstructing dynamic object from a single monocular video under challenging camera setup, for category-agnostic objects.
    \item We leverage LRM to prepare per-frame coarse geometry prior and further propose a deformation MLP to build the correspondence of frames and to faithfully match the dynamic information of the input video.  
    \item We proposed a novel mesh Gaussian structure that provides higher fidelity appearance, lower memory burden, and higher training speed. We also proposed two constraints for mesh Gaussians to avoid view-overfitting.
\end{itemize}

\section{Related Work}
\textbf{Differentiable Rendering} NeRF~\cite{lin2022enerf} was proposed to represent a static scene with density and color volumes defined by an MLP. The training process is time-consuming as it performs costly numerical integration along camera rays at every training iteration. Recently, a more efficient method, 3D Gaussian Splatting~\cite{kerbl2023_3dgs} was proposed. It represents a scene as many anisotropic 3D Gaussian volumes with tractable integrations. It achieved great speedup compared to NeRF. However, due to the anisotropic nature of 3D Gaussians, they are prone to overfitting to camera views, often elongating along the camera ray. This results in artifacts when rendered from novel viewpoints, particularly in sparse-view regions where multi-view supervision is weak. Scaffold GS~\cite{lu2024scaffold} partially solved the overfitting issue by encoding local geometric structures in compact neural features. 2D GS~\cite{huang20242d} proposed to set one of the axis of 3D GS to have almost zero scale in order to model geometrical details more accurately. Building on top of that, mesh based GS~\cite{guedon2024sugar, gao2024mesh} anchor Gaussian Splats to mesh faces, with the Gaussian normal direction aligned with the mesh faces either through a loss term or a hard constraint. An underlying geometric representation makes them less susceptible to overfitting. Although they thrive in multi-view reconstruction tasks, they still suffer from view-overfitting in monocular reconstruction tasks, even worse on dynamic scenes.

\textbf{Dynamic Reconstruction}
Recent works have used both NeRF and 3D GS for reconstructing dynamic scenes. NeRF based methods~\cite{fang2022TiNeuVox, park2021hypernerf, pumarola2021d, park2021nerfies, cao2023hexplane, li2022neural} usually uses a time-conditioned or per-frame embedding conditioned NeRF to model time-varying appearance of scenes. 3D GS based methods~\cite{luiten2023dynamic, wu20244d, yang2024deformable} also uses time-conditioned deformation models to offset a canonical set of 3D GS points to respective timesteps to match the captured images. These works have all achieved incredible results on existing monocular video dataset of dynamic objects.

Nonetheless, most of these methods are not robust against novel-view rendering, and most existing datasets either provide effectively multi-view input or use testing views that's not too far from the training view, as pointed out by~\cite{gao2022monocular}. There are two levels of difficulties here: 
\begin{itemize}
    \item \textbf{Lack of geometric grounding.} Without an explicit geometric representation, photometric appearances remain unconstrained, reducing the reconstruction to mere colored points in space.
    \item \textbf{Weak coupling between geometry and appearance.} Even when a geometric representation is present, existing appearance models are not strictly constrained by it, leading to potential view-overfitting.
\end{itemize}
As a result, most successful and deployable monocular reconstruction methods either rely on class-specific templates (e.g., SMPL)~\cite{peng2021neural-body, shao2022doublefield, weng2022humannerf, jiang2022neuman, suo2021neuralhumanfvv, shao2022floren, kwon2021nhp, pan2023transhuman} or require additional depth input to compensate for missing multi-view information.

\section{Method}
\begin{figure*}
    \centering
    \includegraphics[width=\linewidth]{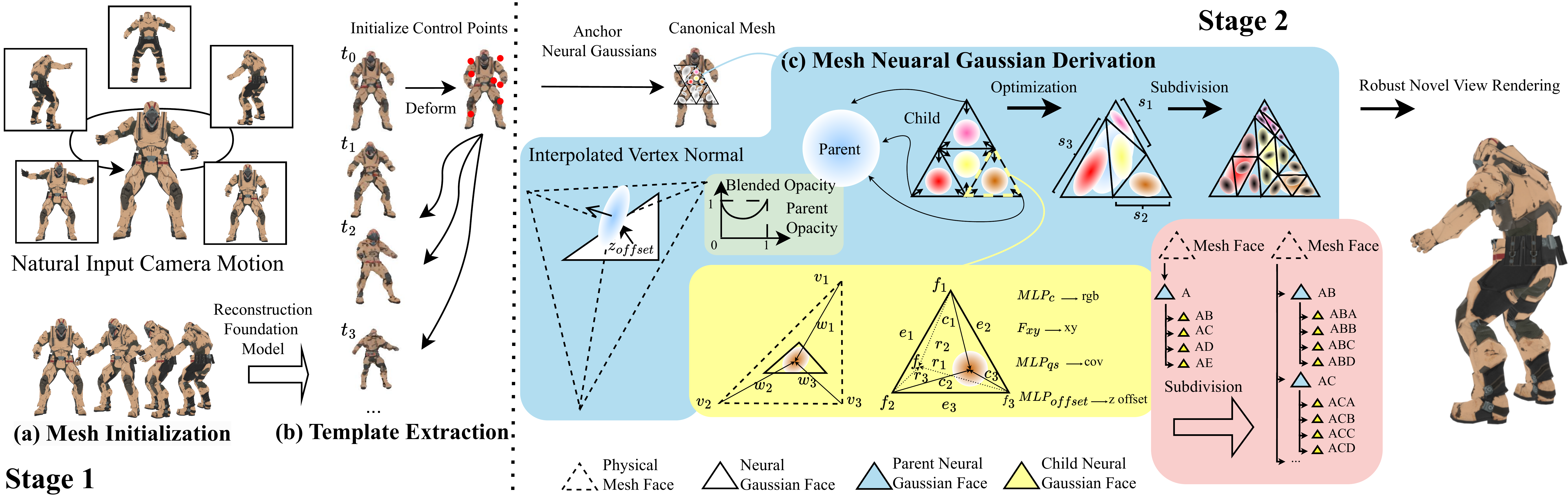}
    \caption{\textbf{Illustration of pipeline.} In stage one, we get per-frame mesh sequence from LRM by querying each frame. We fix the mesh by Taubin smoothing~\cite{taubin} and subdivide faces or collapse edges until the number of faces reaches our desired initial number of Gaussians. In stage two, we initialize 2D Gaussians defined by neural features on the canonical mesh. We train the neural Gaussians jointly with the deformation model. The resulting Gaussians are extremely robust to view-overfitting.}
    \label{fig:pipeline}
\end{figure*}
Reconstructing both geometry and appearance of objects from slowly moving cameras is challenging due to the ambiguity in motion of unobserved region, view overfitting tendencies of differentiable rendering methods, and lack of geometric information from camera movement. Our model is able to solve the challenge by performing a 2-stage optimization process. In the first stage shown in \cref{fig:pipeline} (a) and (b), we designed a robust framework to extract motion and geometry information from unstable LRM output defined by a canonical mesh and a control-point based deformation model. In the second stage in \cref{fig:pipeline} (c), structured 2D GS will be initialized on the canonical mesh. We express 2D GS as functions of vertex neural features to keep an expressive and compact representation. Robustness to view-overfitting is achieved by avoiding GS occlusion and constraining GS influence to local mesh faces. We also include a carefully designed adaptive GS subdivison mechanism to automatically add new GS to regions with fine photometric or geometric details through a mesh-Gaussian quad tree.
\begin{figure}
    \centering
    \includegraphics[width=\linewidth]{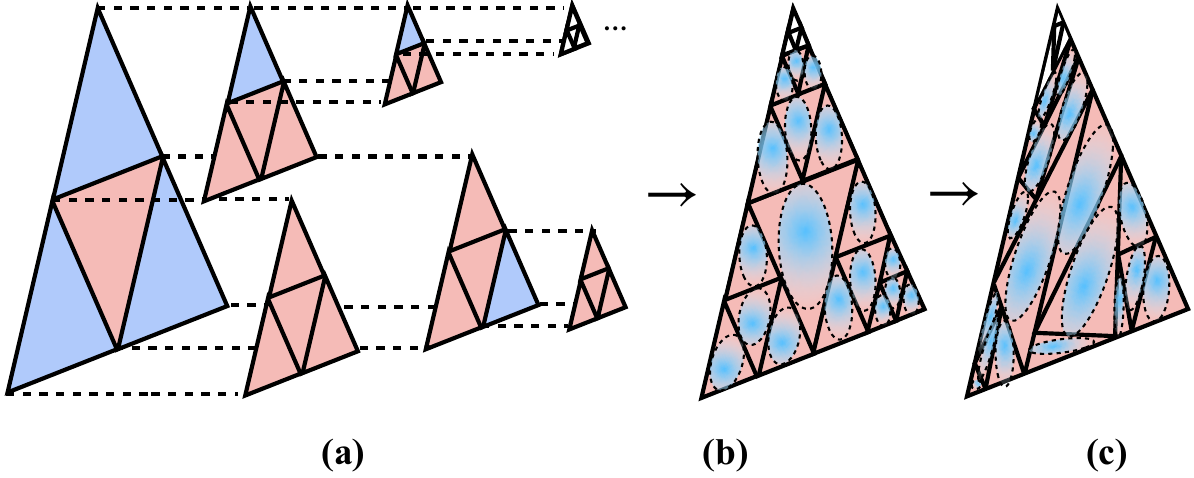}
    \caption{\textbf{Adaptive densification via mesh-Gaussian quad tree on a single mesh face.} Red triangles are leaf nodes of whose associated Gaussians will not further subdivide. Blue triangles are non-leaf nodes with no associated Gaussians. (a) and (b): the tree allows for adaptive density of Gaussians. (c): learnable subdivision ratio further improves the expressiveness.}
    \vspace{-3mm}
    \label{fig:quad}
\end{figure}
\subsection{Stage One: Data Driven Template}

In the first stage, we extract 3D template and motion information from LRM~\cite{tochilkin2024triposr, honglrm} to form a 4D template.
We have LRM output corresponding meshes to each frame of the input video and choose one (in our experiments simply the first mesh) to be canonical mesh. We then perform ICP (Iterative Closest Point) between canonical mesh and each of the meshes in the sequence. Then similar to BANMo~\cite{yang2022banmo}, our deformation model initializes control points and optimize their positions and a time-conditioned MLP to drive their motions. During this process, we jointly optimize the skinning weight of mesh vertices to drive the mesh.
\begin{equation}
    \begin{aligned}
    w_{n,k}&=\text{MLP}_{weight,k}(\mathbf{C}_{k}-\mathbf{v}_{n,t_0})+{\Large \phi}(||\mathbf{C}_{k}-\mathbf{v}_{n,t_0}||)\\
    \mathbf{c}_{k,t}&=\text{MLP}_{motion,k}(t), \ \mathbf{v}_{n, t} = \mathbf{v}_{n,t_0} + \sum^K_{k=1}{\mathbf{c}_{k,t}w_{n,k}}
    \end{aligned}
\end{equation}
In the above equation where we derived the final per vertex location $\mathbf{v}_{n, t}$ at time $t$, $\text{MLP}_{weight,k}$ is a per control point MLP that output weights taking as input displacement from control point to canonical vertices, $\phi$ is an isotropic Gaussian kernel for radial basis function whose scales are determined by the average nearest neighbor distance of the control points at initialization, $w_{n,k}$ is the composite skinning weight between the n\textsuperscript{th} vertex $\mathbf{v}_{n,t_0}$ and the k\textsuperscript{th} control point $\mathbf{C}_k$. $\text{MLP}_{motion,k}$ is a per control point MLP that takes as input time and output control point displacement $\mathbf{c}_{k,t}$, and $\mathbf{v}_{n,t_0}$ is per vertex location at canonical time.
\begin{equation}
\begin{aligned}
    \mathcal{L}_{RCD} &= \frac{1}{|V|} \sum_{v \in V} \min \left( d^2, \min_{u \in U} \| v - u \|^2 \right)\\
    &+ \frac{1}{|U|} \sum_{u \in U} \min \left( d^2, \min_{v \in V} \| u - v \|^2 \right)\\
    \mathcal{L}_{total}&=w_{lap}\mathcal{L}_{lap} + w_{n}\mathcal{L}_{n}+\mathcal{L}_{RCD}
\end{aligned}
\label{eq:loss_nricp}
\end{equation}
In the above equation, we proposed robust Chamfer loss $\mathcal{L}_{RCD}$, where $U$ and $V$ are sets of vertices in target and source meshes, $d$ is truncation distance, $\mathcal{L}_{lap}$ is mesh Laplacian regularization, $\mathcal{L}_{n}$ is normal consistency regularization, and $\mathcal{L}_{RCD}$ is our proposed robust Chamfer loss; $w_{lap}$ and $w_{n}$ are weights for respective loss terms. We optimize the deformation model against $\mathcal{L}_{total}$. Please refer to the supplementary material for more detail about stage one.
\subsection{Stage Two: Tessellation Gaussian Splatting}
\label{sec:TGS}
In the second stage, we solely use the ground-truth frames to jointly optimize motion information and appearance defined by GS. Traditional GS doesn't perform well when trained on sparse views as they could overfit to training cameras. We have identified 2 major reasons. First, Gaussians could elongate along the camera rays freely without hurting photometric performance. In addition, when large scale deformation happens on non-visible regions from training views, unsupervised occluded Gaussians could appear due to the separation of the occluding Gaussians, resulting in artifacts. We show the examples of these two types of artifacts in our ablation study shown in \cref{fig:locality}. We therefore propose a novel 2D GS architecture, mesh-Gaussian quad tree, that both avoids Gaussian occlusion and scale overfitting by constraining their locations and scales to local structured triangles on mesh faces to minimize Gaussian overlapping.
\vspace{-3mm}
\subsubsection{Structured 2D Gaussians}
We build mesh-Gaussian quad tree as demonstrated in \cref{fig:pipeline} (c) and \cref{fig:quad}. At initialization, each mesh face is also a parent Gaussian face. Each parent face is divided into 4 smaller child faces connecting its 3 edge centers similar to Mesh-GS~\cite{gao2024mesh}, forming a quad tree shown in \cref{fig:quad} (a) and (b). However, instead of fixing the subdivision points like MeshGS, we introduce learnable edge point ratios $s_1$, $s_2$, and $s_3$ to allow these points to slide along the edges to match local textures, as in \cref{fig:quad} (c). A large parent Gaussian is spawned at the center of the parent face, while 4 child Gaussians are spawned at the center of child faces. To maintain clarity, we refer to both parent faces and child faces as Gaussian faces, distinguishing them from the underlying physical mesh faces, whose structure remains unchanged throughout the training. This hierarchical structure enables adaptive refinement of Gaussians, resulting in a compact and expressive representation.\\
Each parent face has 6 learnable features $f_i\in\mathbb{R}^{128}$: 3 parent features are defined on the vertices, and 3 edge features on the edges. Each Gaussian has a learnable barycentric weight $\mathbf{r}\in\mathbb{R}^3$ to interpolate from the 3 vertices to decide their barycentric coordinate. They have a separate set of barycentric weights $\mathbf{c}\in\mathbb{R}^3$ to interpolate from the 3 vertex features to get their neural features. For both parent and child Gaussians, features are interpolated as follows:
\begin{equation}
    \label{eq:interpolate}
    f_{interpolated}=\text{softmax}(\mathbf{r})\cdot [f_1||f_2||f_3]
\end{equation}
where $f_1$, $f_2$, and $f_3$ are vertex features of the faces each Gaussian belong to, and $||$ is the concatenation operator. For parent faces, vertex features are directly stored. For child faces, for example the yellow triangular face in \cref{fig:pipeline}, its vertex features are calculated as:
\begin{equation}
    \begin{aligned}
    \label{eq:feature}
    f_1=f_{parent_1}\cdot(1&-s_1)+f_{parent_3}\cdot s_1 + f_{edge_2}\\
    f_2=f_{parent_2}\cdot s_2&+f_{parent_3}\cdot (1-s_2) + f_{edge_3}
    \end{aligned}
\end{equation}
and $f_3$ equals $f_{parent_{3}}$. We interpolate parent vertex features $f_{parent}$ along edges by our learnable shape proportion $s$ and add the edge features $f_{edge_i}$ on top of that. We allow child faces to share vertex features with parent face where they share a common vertex. We decode Gaussian features with 3 MLP appearance decoders:
\begin{equation}
\begin{aligned}
\label{eq:decode}
\text{MLP}_{qs}([f_{interpolated}||\frac{e2}{e1}||\frac{e3}{e1}])&=[\mathbf{q_{_{2D}}}||\mathbf{s_{_{2D}}}]\\
\text{MLP}_{c}([f||\mathbf{p}])&=rgb\\
\text{MLP}_{offset}(f_{interpolated})&=z_{offset}
\end{aligned}
\end{equation}
where $\mathbf{q_{_{2D}}}||\mathbf{s_{_{2D}}}$ is the 2D GS's rotation and scale concatenated. $z_{offset}$ is the GS offset along the face normal direction, $e_1$, $e_2$, and $e_3$ are the edge lengths of each Gaussian face. To model pose dependent apperance, we encode the location of the 30 control points into a pose embedding $\mathbf{p}$, similar to Gaussian Avatar~\cite{hu2024gaussianavatar} and Animatable Gaussians~\cite{li2024animatable}. 2D GS's normal direction is interpolated from vertex normals, and we similarly interpolate vertex colors to add to the decoded neural colors. Further GS constraints are described in \cref{sec:constraints}.
\subsubsection{Competitive Gaussian Opacities}
\begin{figure}
    \centering
    \includegraphics[width=\linewidth]{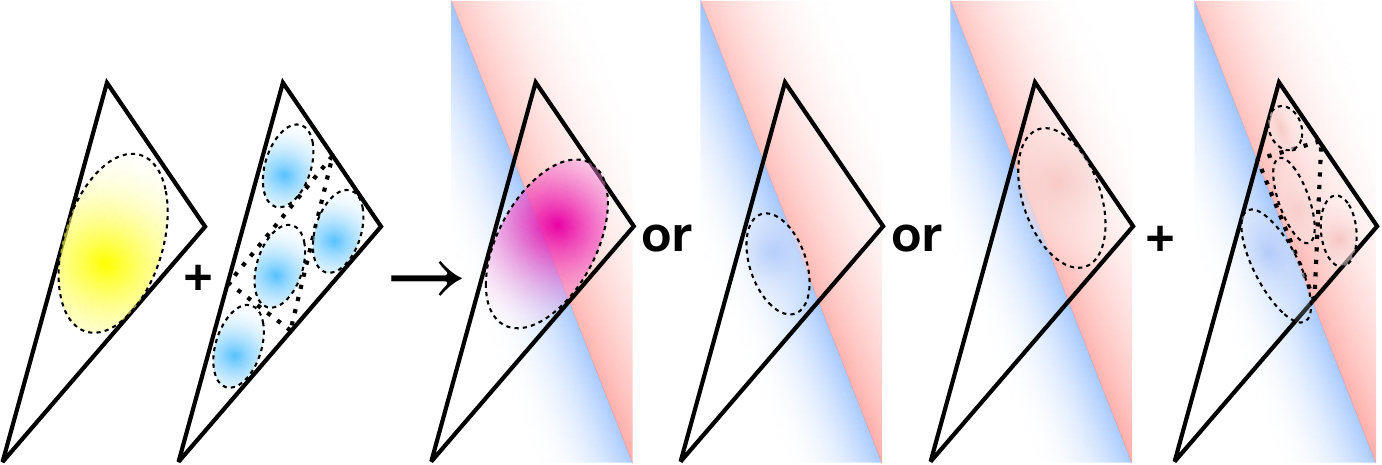}
    \caption{\textbf{Learnable subdivision ratio fits boundary better.} Yellow Gaussian is a parent Gaussian, blue Gaussians are child Gaussians. Color boundary denoted by red and blue regions can be better modeled by child Gaussians than parent. Child Gaussians' opacities will naturally become one through optimization.}
    \vspace{-3mm}
    \label{fig:subdiv}
\end{figure}

We utilized Gaussian opacities for gradient-free Gaussian subdivision. Specifically, only parent Gaussians have optimizable opacities as in \cref{fig:subdiv}, while the four child Gaussians compete their opacities with their parents, as computed in \cref{eq:opacities}, where we have heuristically set $\beta$ to be 0.9. Importantly, $\beta$ ensures the sum of a parent Gaussian’s opacity and any of its child Gaussians' opacities never equals 1 except when the parent Gaussian's opacity is either 0 or 1, as plotted in \cref{fig:pipeline}. In such cases, the child Gaussians’ opacities become either 0 or 1, effectively forcing binary opacity assignments.
\begin{equation}
    \label{eq:opacities}
    \alpha_{child} = (1-\alpha_{parent}^\beta)^{\frac{1}{\beta}}
\end{equation}
With this setup, each parent Gaussian's opacity serves as an indicator of local detail level. The regularization term $\mathcal{L}_{\alpha}$ in \cref{eq:alpha}, which reaches its minimum only when all parent Gaussians are fully opaque, competes with the photometric and geometric loss terms. When the other loss terms exceed a certain threshold, indicating the presence of finer local details, parent Gaussians’ opacities tend to zero, allowing child Gaussians to emerge and model these finer details. This mechanism adaptively allocates Gaussian population, ensuring that regions with higher complexity receive higher-resolution representations, while simpler areas remain efficiently represented by fewer and larger parent Gaussians, thereby maintaining an adaptive and efficient Gaussian hierarchy.
\subsubsection{Gaussian Constraints}
\label{sec:constraints}
We propose two constraints on the Gaussians to avoid view-overfitting. First, Gaussian scales are constrained to a maximum of one-fourth of the base and height of their respective triangles by applying a sigmoid activation to the decoded $\mathbf{s_{2D}}$ in \cref{eq:decode}. Gaussians are also allowed to rotate around normals by the decoded rotation offset $\mathbf{q_{2D}}$. This way, the Gaussians are initialized with just enough scale to span the surface which naturally minimizes overlapping. As the subdivision mechanism progresses, new Gaussians are dynamically introduced to fill regions where the initial Gaussians fail to provide sufficient coverage.

In addition, for Gaussian offset along the face normal direction, rather than employing a soft Gaussian anchoring constraint as in DG-Mesh~\cite{liu2024dynamic}, we introduce a scale-dependent anchoring strategy. Gaussians associated with smaller neural faces are permitted to have larger offsets while initial Gaussians with parent being mesh faces have zero offset, judged by the value:
\begin{equation}
    u=\tanh(\frac{e_p}{e_g})\\
\end{equation}
where $e_{p}$ and $e_{g}$ are the mean edge length of the root mesh face and the mean edge length of the Gaussian face. However, no offset is allowed to exceed the mean edge length of the root physical mesh face, ensured by
\begin{equation}
    \begin{aligned}
    \label{eq:offset}
    w_{bar}=(1-w_1)(1-w_2)(1-w_3)\\
    offset=w_{bar}\cdot u\cdot e_{p} \cdot\tanh(z_{offset})
    \end{aligned}
\end{equation}
where $w_1$, $w_2$, and $w_3$ are the barycentric coordinates of Gaussian points with respect to their root mesh faces.

This ensures that finer geometric details are only carved after the object’s motion has already been well optimized, which prevents premature deformations that could lead to inconsistent geometry. This constraint also ensures that strong gradient will flow to mesh vertices to optimize geometric details through photometric losses. The effectiveness of the contraints are shown in \cref{sec:ablation}.
\subsubsection{Adaptive Gaussian Population Control}
\label{sec:adaptive pop}
Thanks to our proposed adaptive mesh subdivision process, our neural GS can model very fine level of details with our mesh-Gaussian quad tree, as illustrated by \cref{fig:pipeline}. Our model subdivides a parent Gaussian by turning its 4 child Gaussians into 4 new parent Gaussians, making each child face a new parent face. During this process, we remove the original parent Gaussian, compute its three edge features, and reassign them as new vertex features. The newly introduced edge features are initialized to zero, minimizing disruption to the optimization process and allowing us to constantly introduce new Gaussians in a stable manner during training. Edge features of each new parent faces are initialized to zero, minimizing the impact on optimization process and incrementally introducing new Gaussians in a stable manner during training. Our subdivision mechanism is implemented as follows:
\begin{itemize}
    \item \textbf{Parent Gaussian Subdivision:} All parent Gaussians with opacities below 0.1 are subdivided every 5000 iterations except for the initial and last 5000 iterations, ensuring denser Gaussians in finer regions.
    \item \textbf{Child Gaussian Deactivation:} Excessive child Gaussians whose parent opacities remain above 0.9 in 90\% of the iterations since the last subdivision are turned off, and we also exclude them when calculating opacity regularization $\mathcal{L}_{\alpha}$.
\end{itemize}
\subsubsection{Loss Terms}
Our losses are as follows:
\begin{equation}
    \begin{aligned}
        \mathcal{L}_{pho}&=\mathcal{L}_{1}+\mathcal{L}_{ssim}
    \end{aligned}
\end{equation}
where we used $\mathcal{L}_1$ and SSIM loss between GT image and rendered image to supervise photometric appearance,
\begin{equation}
    \mathcal{L}_{edge}=\frac{1}{|\mathcal{E}|} \sum_{(i,j) \in \mathcal{E}} (\|\mathbf{v}_{i, t_0} - \mathbf{v}_{j,t_0}\|-\|\mathbf{v}_{i, t} - \mathbf{v}_{j,t}\|)^2
\end{equation}
where we penalize edge length changes with respect to canonical mesh, $\mathbf{v}_{i, t_0}$ is the canonical vertex coordinate and $\mathbf{v}_{i, t}$ is its coordinate at timestep $t$. $\mathcal{E}$ is the set of neighboring vertices.
\begin{equation}
    \label{eq:alpha}
    \mathcal{L}_{\alpha}=\frac{1}{N}\sum_{i}^{N}\alpha_{i}
\end{equation}
The above $\mathcal{L}_{\alpha}$ is the mean opacity of all Gaussians. Due to our parent-child opacity coupling, its minimum is when all parents are fully opaque.
Our final loss $\mathcal{L}_{TGS}$ is:
\begin{equation}
    \begin{aligned}
    \label{eq:loss_tgs}
    &\mathcal{L}_{reg}=\mathcal{L}_{lap}+w_{edge}\mathcal{L}_{edge}+w_{\alpha}\mathcal{L}_{\alpha}\\
    &\mathcal{L}_{prior}=w_{flow}\mathcal{L}_{flow}+w_{normal}\mathcal{L}_{normal}\\
    &\mathcal{L}_{TGS} = \mathcal{L}_{pho}+w_{reg}\mathcal{L}_{reg}+w_{prior}\mathcal{L}_{prior}
    \end{aligned}
\end{equation}
where $\mathcal{L}_{flow}$ is the difference between the rendered optical flow and the optical flow predicted by RAFT~\cite{teed2020raft}. $\mathcal{L}_{lap}$ is mesh Laplacian regularization further explained in supplementary material. We use normal supervision $\mathcal{L}_{normal}$ provided by DSINE~\cite{bae2024rethinking} only for static cameras.

\begin{figure*}
    \centering
    \includegraphics[width=\linewidth]{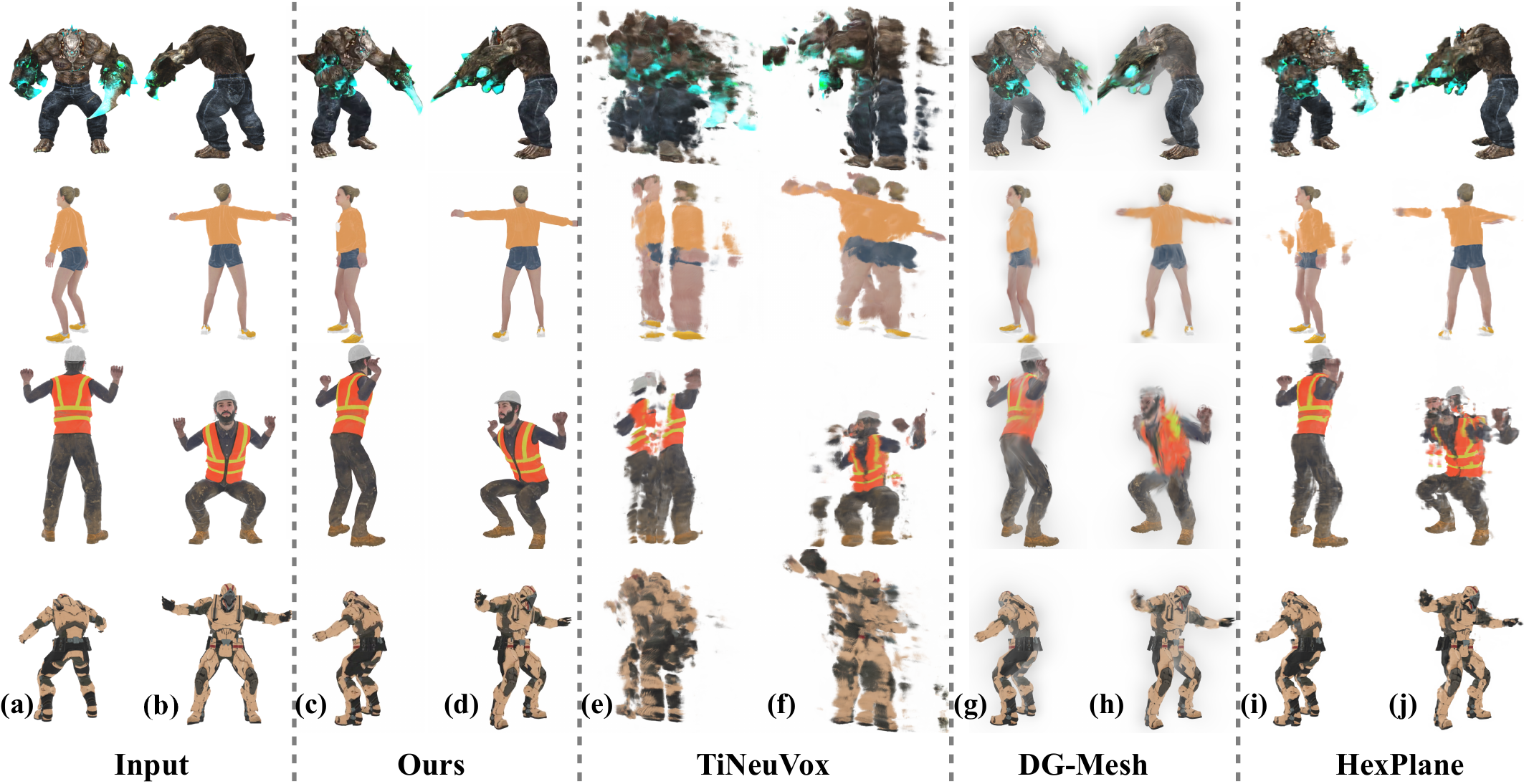}
    \caption{\textbf{Test results on Smooth D-NeRF.} (a) and (b): input images at two different timesteps. (c), (e), (g), and (i): rendering results at the first timestep. (d), (f), (h), and (j): rendering results at the second timestep. Our results are visually better than all other methods. The second best results are produced by DG-Mesh~\cite{liu2024dynamic}. Their 3D Gaussians' unconstrained scales result in foggy appearance caused by large Gaussian floaters.}
    \vspace{-3mm}
    \label{fig:visual}
\end{figure*}
\begin{figure}
    \centering
    \includegraphics[width=\linewidth]{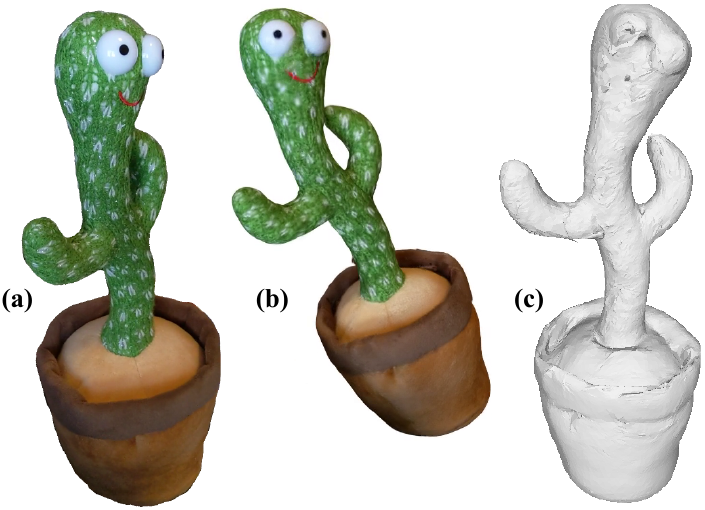}
    \caption{\textbf{Unbiased4d~\cite{johnson2023unbiased} results.} (a): input image. (b): rendered novel view. (c): extracted mesh.}
    \label{fig:ub4d}
    \vspace{-3mm}
\end{figure}
\section{Experiments}
\begin{figure*}
    \centering
    \includegraphics[width=\linewidth]{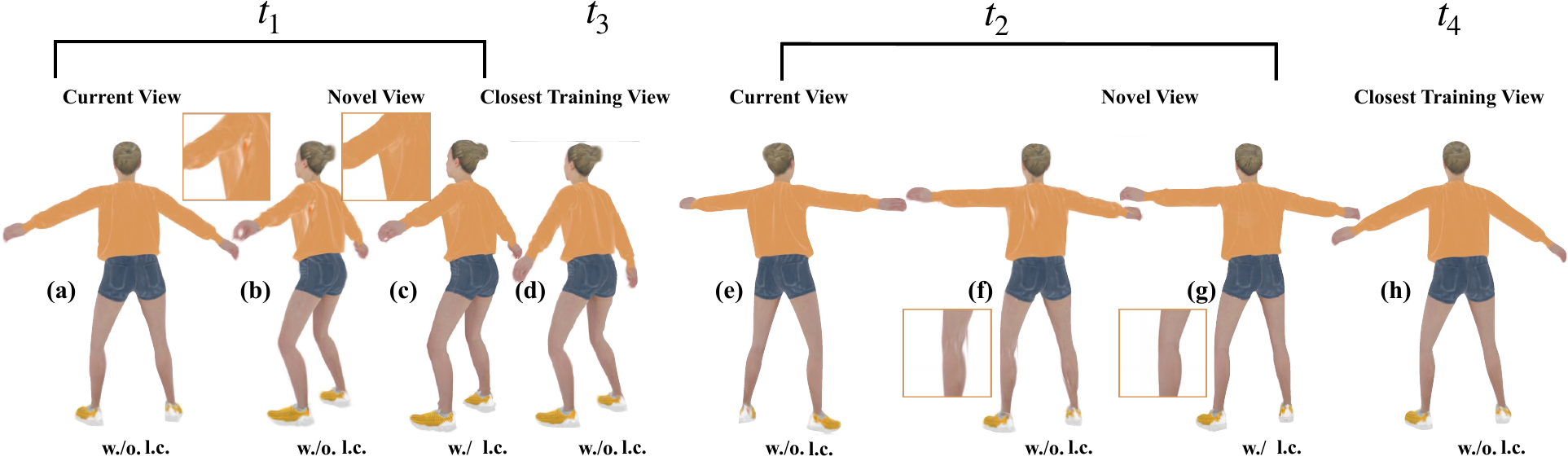}
    \caption{\textbf{Ablation comparison of locality constraints (l.c.).}  (a) and (e): ablation rendering results from training pose at current timestep. (b) and (f): ablation rendering results from test pose. (c) and (g): rendering results from test pose without ablation. (d) and (h): spatially closest training views to test views at $t_1$ and $t_2$. (b) demonstrated view-overfitting caused by occluded Gaussians not being optimized, as shown by the large red Gaussian. Comparison between (f) and (g) shows our pipeline avoids 2D GS scales overfitting to the training view. (a) and (e) shows that these artifacts are not be visible to training views but will appear in novel test views.}
    \label{fig:locality}
\end{figure*}
\begin{table*}[h!]
    \centering
    \label{tab:example}
    \resizebox{0.8\textwidth}{!}{
    \begin{tabular}{ccccccccccc}
    \hline
    \toprule
    \multirow{2}{*}{Method} & &  \multicolumn{4}{c}{Hook}   &  & \multicolumn{4}{c}{Mutant}\\ \cline{3-6} \cline{8-11}
               &&   PSNR $\uparrow$ & SSIM $\uparrow$ & LPIPS $\downarrow$ & CD $\downarrow$ &  & PSNR $\uparrow$ & SSIM $\uparrow$ & LPIPS $\downarrow$ & CD $\downarrow$  \\ \hline
    TiNeuVox-B    && 13.996&0.859&0.1803&-&  &16.380&\bf{0.914}&0.1801&-\\  
    HexPlane    &&27.836&0.966&0.0297&-&  &17.632&0.880&0.1126&-\\
    DG-Mesh &&26.123&0.968&0.0506&2&  &\bf{18.506}&0.879&0.1057&\bf{2.8}\\
    Ours &&\bf{32.302}&\bf{0.981}&\bf{0.0208}&\bf{0.8}&  &17.104&0.882&\bf{0.1017}&3.8\\ 
    \bottomrule
    \toprule
    \multirow{2}{*}{Method} & &  \multicolumn{4}{c}{Standup}   &  & \multicolumn{4}{c}{Jumping Jacks}\\ \cline{3-6} \cline{8-11}
               &&   PSNR $\uparrow$ & SSIM $\uparrow$ & LPIPS $\downarrow$ & CD $\downarrow$ &  & PSNR $\uparrow$ & SSIM $\uparrow$ & LPIPS $\downarrow$ & CD $\downarrow$  \\ \hline
    TiNeuVox-B    && 12.583&0.839&0.2446&-&  &16.417&0.915&0.1799&-\\  
    HexPlane    &&21.256&0.898&0.0936&-&  &24.977&0.940&0.0770&-\\
    DG-Mesh &&22.481&0.937&0.0875&1.1&  &25.433&0.962&0.0665&7\\ 
    Ours &&\bf{23.464}&\bf{0.942}&\bf{0.0572}&\bf{0.9}&  &\bf{25.995}&\bf{0.964}&\bf{0.0404}&\bf{1.1}\\ 
    \bottomrule
    \end{tabular}
}
    \vspace{3mm}
    \vspace{-0.3in}
    \label{tab:main_res}
    \vspace{2mm}
    \caption{Quantitative comparison against previous works on Smooth D-NeRF. The best method in each metric is bolded. Chamfer distance (CD) is reported in scale $1\mathrm{e}{-3}$ and is unitless, same as training data of LRM.}
\end{table*}
\subsection{Datasets}
Original D-NeRF dataset, which employs teleporting cameras to simulate multi-view supervision, is impractical for real-world applications. Thus we re-render D-NeRF’s humanoid characters\footnote{Available on Adobe's Mixamo.} with naturally moving cameras.
We rotate the camera around the character at constant speed and return camera to the starting position as the animation stops. We place 2 test cameras rotating with the training camera, both looking at the character but are 45 degrees away from the training camera. Further illustration of the setup is available in the supplementary material.
In addition to blender rendered dataset, we qualitatively tested our model on Unbiased4D~\cite{johnson2023unbiased}, including one video of a deforming cactus toy captured using a hand-held camera. We also tested on People-Snapshot~\cite{alldieck2018video} comparing against Gaussian Avatar~\cite{hu2024gaussianavatar}, featuring self-rotating actors captured with a static camera. We train our model on a single 3090 GPU with Adam~\cite{kingma2014adam} and PyTorch. The whole training pipeline takes about 90 minutes.
\begin{figure}[h!]
    \centering
    \includegraphics[width=\linewidth]{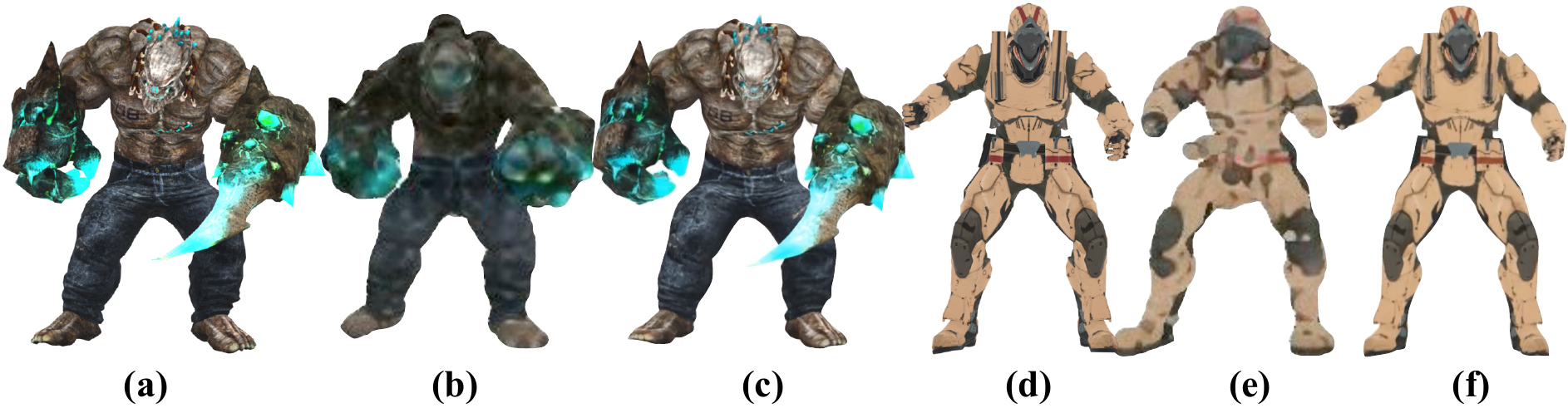}
    \caption{\textbf{Comparison with LRM direct output.} (a) and (d): ground-truth. (b) and (e): LRM direct output. (c) and (f): our pipeline's output. Further comparison available in supplementary material.}
     \vspace{-5mm}
    \label{fig:lrm_comp}
\end{figure}
\subsection{Main Results}
\begin{figure*}
    \centering
    \includegraphics[width=\linewidth]{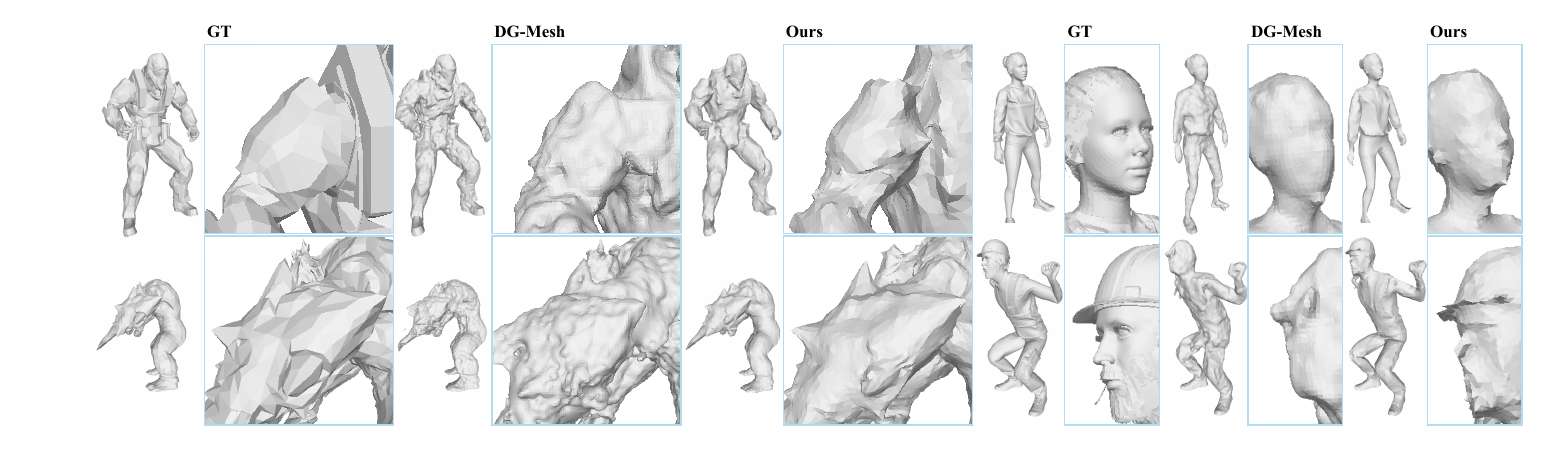}
    \caption{\textbf{Qualitative comparison of mesh reconstruction.} Our meshes contain much more geometric details due to the strong correlation between the photometric appearance and geometries brought by the Gaussian offset constraint.}
    \vspace{-5mm}
    \label{fig:mesh_comp}
\end{figure*}
\begin{figure}
    \centering
    \includegraphics[width=\linewidth]{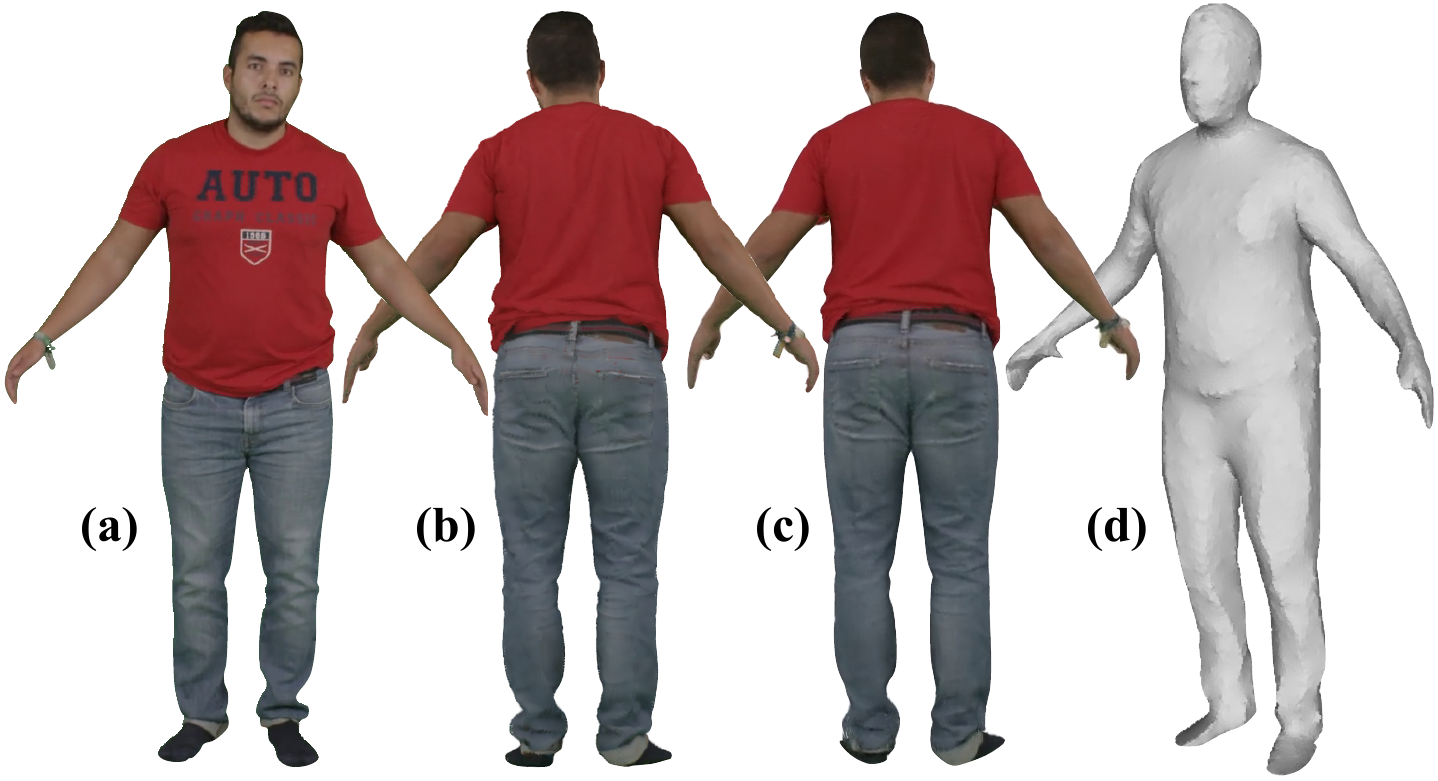}
    \caption{\textbf{People-Snapshot results.} (a): input view rendering result. (b): our method's novel view rendering result. (c): Gaussian Avatar's novel view rendering result. (d): our extracted mesh.}
    \vspace{-3mm}
    \label{fig:pss}
\end{figure}

Our main results on Smooth D-NeRF are demonstrated in \cref{fig:visual}, \cref{fig:mesh_comp}, and \cref{tab:main_res}. We achieved better results than all compared methods. Since TiNeuVox and HexPlane lack an underlying geometric representation, they exhibit severe overfitting to training views. DG-Mesh, while more robust due to mesh-anchored Gaussians, still suffers from mild view-overfitting due to its use of vanilla 3D GS. However, the "mutant" data features a relatively stationary motion, and our model's flexible motion representation is disadvantaged but still achieved better perceptive quality measured by LPIPS. Shown in \cref{fig:mesh_comp}, DG-Mesh generates noisy meshes due to its reliance on SfM to initialize meshes, which performs poorly when object motion dominates over camera motion, and its direct use of time-conditioned MLPs for deformation, which is under-constrained. Our method produces accurate meshes with sharp details both because we initialize meshes directly from 3D priors and our use of merely 30 learnable control points to drive our meshes, providing strong local rigidity guarantee. We also show side-by-side comparison with LRM's direct rendering result in \cref{fig:lrm_comp}, and our model refines both geometry and appearance significantly. In addition, our results shown in \cref{fig:ub4d} and \cref{fig:pss} shows our model is robust against in-the-wild camera motion and static camera, achieving similar performance to Gaussian Avatar~\cite{hu2024gaussianavatar}.
\subsection{Ablation Studies}
\label{sec:ablation}
\subsubsection{Effectiveness of Pruning Strategy}
We tested our pipeline without child Gaussian pruning mechanism. As shown in \cref{tab:pop}, the improvement in PSNR is minimal, suggesting excessive Gaussians that don't contribute much to the photometric quality have indeed be pruned. The number of Gaussians, training time, and GPU memory usage also become much worse without child Gaussian pruning.
\subsubsection{Effectiveness of Locality Constraints}
We allow the 2D Gaussians to freely optimize their attributes, only keeping their normal direction aligned to the interpolated face normal and scale proportional to face size, similar to Mesh-GS~\cite{gao2024mesh}. Shown in \cref{fig:locality}, two types of artifacts occur: occluded Gaussians not being optimized, shown by the large red Gaussian in \cref{fig:locality}(b) that's not visible to any training view, and thin long Gaussians shown in \cref{fig:locality}(f) being Gaussians freely elongate along viewing direction. Our model is able to suppress both artifacts with our scale locality constraints.
\begin{table}[h!]
    \centering
    
    \label{tab:pop}
    \footnotesize 
\setlength{\tabcolsep}{1.8mm}
\begin{tabular}{cccccc}
    \hline
    \toprule
     & PSNR & CD & Num GS & Time & Memory\\ \hline
    w./o. prune & 24.860 & -  & $\sim$750k & $\sim$240 mins & 23GB\\
    w./o. offset & 24.537 & 2.53 & - & - & -\\
    Full model & 24.716 & 1.65 & $\sim$40k & $\sim$80 mins & 6GB\\
    \bottomrule
\end{tabular}
    \caption{Ablation comparison with and without offset constraint and child Gaussian pruning. CD reported in $1\mathrm{e}{-3}$ scale.}
    \vspace{-6mm}
\end{table}
\subsubsection{Effectiveness of Offset Constraints}
\begin{figure}[h!]
    \centering
    \includegraphics[width=\linewidth]{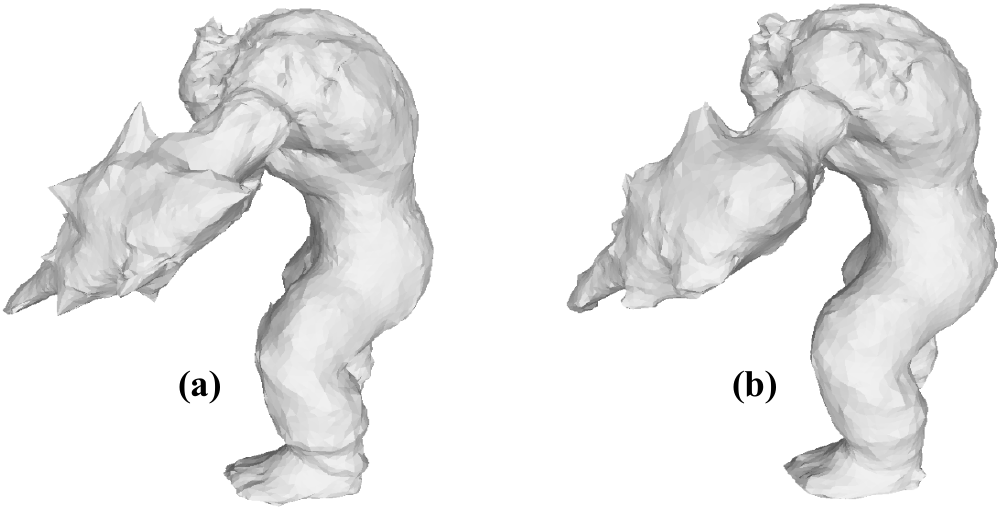}
    \caption{\textbf{Comparison of meshes extracted with and without offset constraint.} (a): mesh with Gaussian offset constraint. (b): mesh without Gaussian offset constraint.}
    \vspace{-3mm}
    \label{fig:offset}
\end{figure}

To test the effectiveness of the offset constraint discussed in \cref{sec:constraints}, we remove the constraint and let the Gaussians freely offset along the face normal. We tested the average chamfer distance of the mesh sequences under this setup, and as shown in \cref{tab:pop}, the chamfer distance is higher. As shown in \cref{fig:offset}, the generated mesh seems coarser and lacks fine details, which suggests our offset constraint indeed allows for more accurate geometric reconstruction from photometric supervision.

\section{Conclusion}
We present Tessellation GS in this work, a pipeline to reconstruct dynamic objects using mesh and novel structured neural Gaussians with robust view extrapolation performance from natural monocular videos. Tessellation GS distributes 2D Gaussians on mesh faces in a structured way with minimal overlapping and decode their attributes from neural features on the vertices. Their population is adaptively controlled through carefully designed mesh-Gaussian quad trees linked by Gaussian opacities that adaptively add more Gaussians in regions with finer details. With scales strongly linked and constrained by mesh face shapes, Tessellation GS has shown excellent robustness to view-overfitting and achieved SOTA performance on monocular reconstruction task using natural camera motion in terms of both photometric performance and mesh quality. 
\vspace{-2mm}
\subsection*{Limitation}
Our method relies on the quality of canonical mesh to handle topological changes. Further studies towards mesh merge and division with consistent mesh face correspondence could consider incorporating Tessellation GS for appearance representation.
\clearpage
{
    \small
    \bibliographystyle{ieeenat_fullname}
    \bibliography{main}
}
\clearpage
\maketitlesupplementary

\section*{Camera Setup}
\vspace{-3mm}
\begin{figure}[h]
    \centering
    \includegraphics[width=\linewidth]{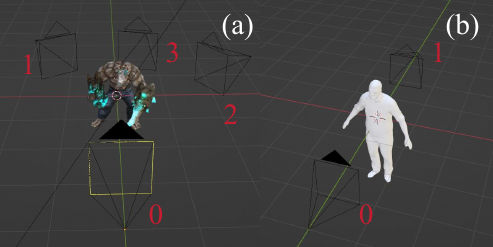}
    \vspace{-20pt}
    \caption{\textbf{Camera setup for SDNF and PSS.}(a): SDNF cameras. (b): PSS cameras.}
    \label{cameras}
\end{figure}
We use blender coordinate system, and the character's up direction aligns with the positive Z-axis.\\
For Smooth D-NeRF (SDNF), shown in \cref{cameras} (a), camera 0 is a static validation view; camera 1, 2, and 3 rotate around the character (z-axis) but are mutually rigid. Camera 3 is training view, and camera 1 and 2 are test views, both 45 degrees azimuthal angles away from camera 3. Video supplementary material's "smooth\_dnerf\_validation"plementary material compares training and validation videos.\\
For People Snapshot dataset (PSS) shown in \cref{cameras} (b), both training camera 0 and test camera 1 are static, and on the opposite sides of the character, 180 degree azimuthal angle to each others. We solely used video captured by camera 0 as input, and tested the results with camera 1. For the one cactus video from Unbiased4D (Ub4D), we could not provide rendering result from similarly novel views since only the front of the toy is captured in the training video.
\vspace{-2mm}
\section*{Stage One Loss Setup}

For our proposed robust Chamfer loss $\mathcal{L}_{RCD}$ in \cref{eq:loss_nricp}, $U$ and $V$ are sets of vertices in target and source meshes, and $d$ is truncation distance. This setup helps the deformation model to not be interfered by large and sudden deformations between consecutive meshes, which are usually caused by temporal inconsistency and floating artifact of LRM. In addition, we define $\mathcal{L}_{lap}$ as follows:
\begin{equation}
\label{eq:lap}
    \mathcal{L}_{lap} = \frac{1}{N} \sum_{i=1}^N \|\mathbf{v}_i - \frac{1}{|\mathcal{E}_{i}|}\sum_{j\in\mathcal{E}_{i}}\mathbf{v}_j\|^2 
\end{equation}
where $\mathbf{v}_i$ is location of the i\textsuperscript{th} vertex and $\mathcal{E}_{i}$ is the set of neighboring vertices of the i\textsuperscript{th} vertex.
$\mathcal{L}_{n}$ is normal consistency regularization defined as follows:
\begin{equation}
\label{eq:normal_consistency}
    \mathcal{L}_{\text{n}} = \frac{1}{|\mathcal{E}|} \sum_{(i,j) \in \mathcal{E}} \|\mathbf{n}_i - \mathbf{n}_j\|
\end{equation}
where $n_i$ is vertex normal vector of the i\textsuperscript{th} vertex and $\mathcal{E}$ is the set of all neighboring vertex pairs.

\begin{figure}[h!]
    \centering
    \includegraphics[width=\linewidth]{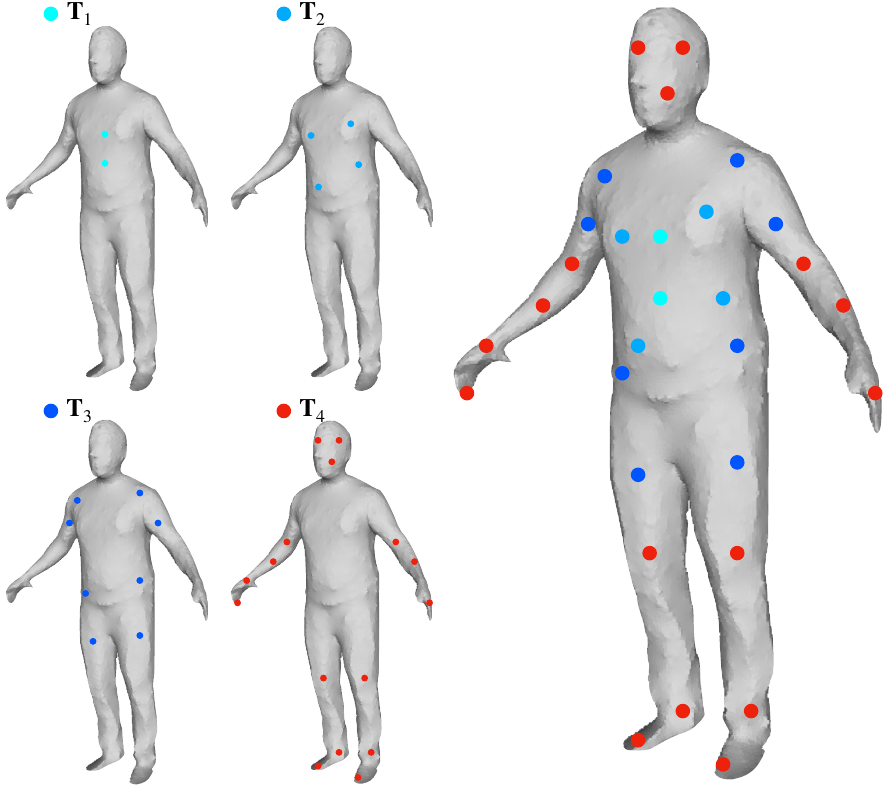}
    \caption{\textbf{Illustration of distribution of control points.} From $T_1$ to $T_4$, each next temperature value is halved with $T_4$ being 1.}
    \vspace{-5mm}
    \label{fig:softmax_struct}
\end{figure}
\begin{table*}[h!]
    \centering
    \resizebox{\textwidth}{!}{
\begin{tabular}{c c c l} 
\hline
 &Loss Term & Weight & Explanation \\ \cline{1-4}
\multirow{3}{*}{Stage One} & $\mathcal{L}_{RCD}$ & 1 & Our proposed robust Chamfer distance that truncate loss to zero above a threshold. \\ \cline{2-4}
                     & $\mathcal{L}_{lap}$ & 0.5 & Mesh Laplacian regularization ensures smoothness of mesh surface and uniformity of vertices. \\ \cline{2-4} 
                     & $\mathcal{L}_{n}$ & 0.001 & Normal consistency regularization ensures smoothness of mesh surface. \\ \hline

\multirow{7}{*}{Stage Two} & $\mathcal{L}_1$ & 0.8 & Rendered frames vs. input frames in L1 norm. \\ \cline{2-4}
                     & $\mathcal{L}_{ssim}$ & 0.2 & SSIM loss between rendered frames and input frames. \\ \cline{2-4}
                     & $\mathcal{L}_{edge}$ & 0.2 & L1 norm loss to penalize change in edge length of meshes after deformation. \\ \cline{2-4}
                     & $\mathcal{L}_{lap}$ & 0.03 & Same as in stage one. \\ \cline{2-4}
                     & $\mathcal{L}_{\alpha}$ & 0.002 & Mean opacity of all Gaussians to encourage the model to use fewer Gaussians. \\ \cline{2-4}
                 & $\mathcal{L}_{normal}$ & 0.1 or 0 & Predicted normal by DSINE~\cite{bae2024rethinking} vs. rendered normal in L2 norm. Only used for static camera cases.\\ \cline{2-4}
                     & $\mathcal{L}_{flow}$ & 0.01 & Predicted optical flow by RAFT~\cite{teed2020raft} vs. rendered optical flow in L1 norm. \\ \hline
\end{tabular}}
    \vspace{-5mm}
    \caption{\textbf{Parameter weights for stage one and stage two.}}
    \vspace{-5mm}
    \label{tab:parameters}
\end{table*}
\vspace{-2mm}
\section*{Deformation Model Setup}
NPGS~\cite{das2024neural} uses a set of low-rank deformation basis to deform points. A fixed set of K learnable deformation basis is used for all timesteps, while each point has an individual learnable time-varying weight used to weighted sum the K deformation basis to get the displacement at timestep $t$. We observed that although it keeps the motion of the object low-rank, it doesn't inherently keep local rigidity. Inspired by DynoSurf~\cite{yao2024dynosurf}, we decided to make things opposite: K deformation control points provide time-varying deformation to drive deformation of mesh vertices.
We express locations of each control point $\mathbf{C}_k$ as weighted sum over all vertices in the canonical mesh with learnable weights $\mathbf{m}_k$ as in \cref{eq:control_weights}, where $\mathbf{V}_{t_0}$ is a matrix whose columns are positions of all vertices in the canonical mesh, this keeps the control points within the convex hull of the mesh.
\begin{equation}
\begin{aligned}
\label{eq:control_weights}
    \mathbf{C}_k&=\mathbf{V}_{t_0}\text{softmax}(\frac{\mathbf{m}_{k}}{T})\\
    &=\sum_{i=1}^{N} \left( \frac{e^\frac{m_{k,i}}{T}}{\sum_{j=1}^{N} e^\frac{m_{k,j}}{T}} \right) \mathbf{v}_{i, t_0}
\end{aligned}
\end{equation}
We used hierarchical temperatures in softmax when deciding control point locations. High temperatures keep the control points close to the large-scale structures of the mesh, while low temperatures allow control points to spread out onto finer geometries of the mesh, analogous to SMPL's tree structure. As illustrated in \cref{fig:softmax_struct}, we use 4 levels of granularities of control points, composed of in total 30 control points to drive the canonical mesh. To initialize the control nodes, we use farthest point sampling (FPS) to select 30 points, then we set the corresponding weight $m_k$ to a large value to keep the result from FPS. Each vertex has a skinning weight over the 30 control points predicted by per control point MLP taking as input displacement from vertex to control point.
\section*{Further Results}
\begin{figure}[h!]
    \centering
    \includegraphics[width=\linewidth]{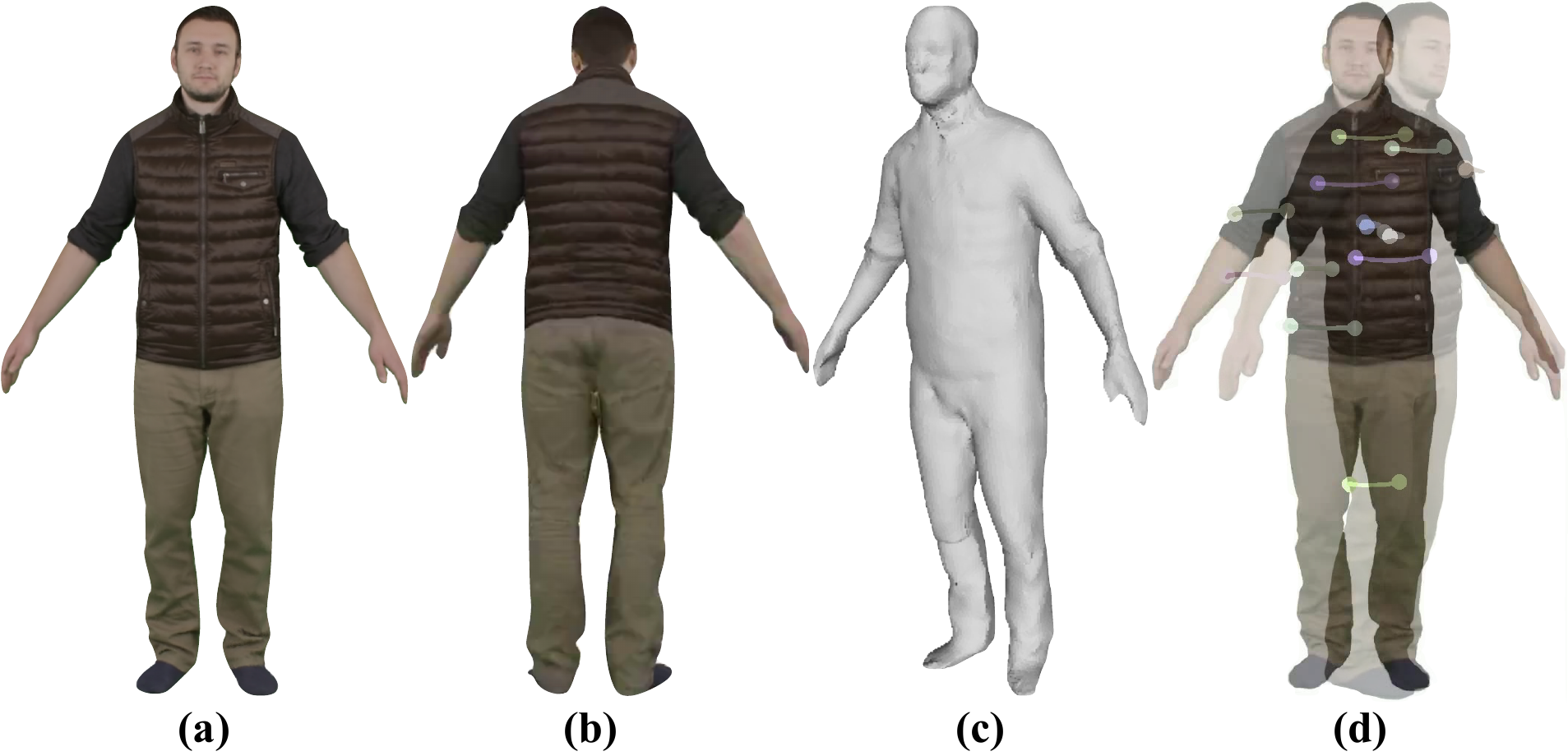}
    \vspace{-3mm}
    \caption{\textbf{People Snapshot result.} (a): input image. (b): novel view rendering result. (c): extracted mesh. (d): tracking result.}
    \label{fig:pss_tracking}
    \vspace{-3mm}
\end{figure}
We have included an additional result on People Snapshot in \cref{fig:pss_tracking}. For complete tracking result, please refer to video results.
\begin{figure}[h!]
    \centering
    \includegraphics[width=\linewidth]{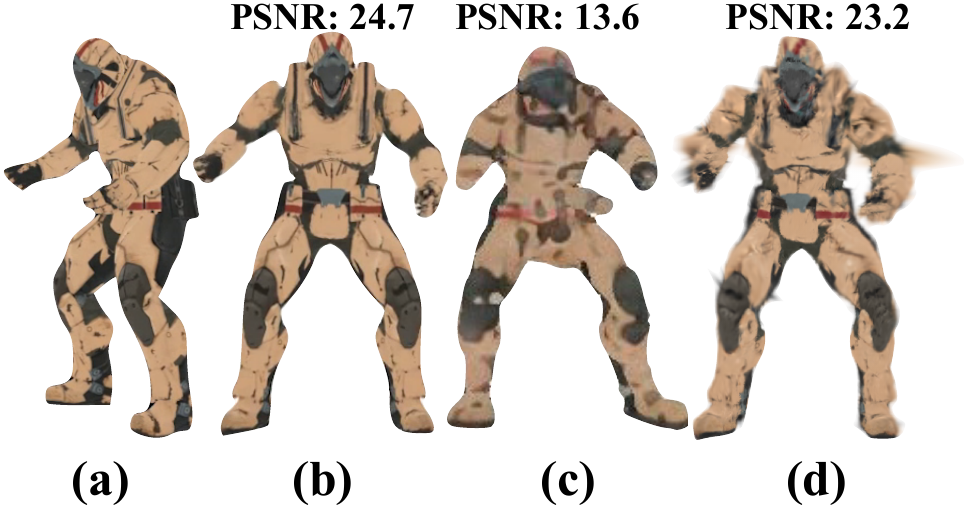}
    \vspace{-3mm}
    \caption{\textbf{Ablation and quantitative comparison on Tessellation GS.} (a): input image. (b): our method's rendered novel view. (c): direct LRM rendering result. (d): DG-Mesh's result after incorporating LRM prior.}
    \vspace{-3mm}
    \label{fig:ablate_lrm}
\end{figure}
We have also conducted further ablation study on the effectiveness of our stage two pipeline. We incorporated stage one of our pipeline with DG-Mesh~\cite{liu2024dynamic}, equipping DG-Mesh with LRM prior. Shown in~\cref{fig:ablate_lrm} (d), DG-Mesh still exhibits view-overfitting due to their relatively more free setup of mesh Gaussians. This proves that both stage one and stage two are crucial for the success of our pipeline.
\section*{Optimization}
In stage one, we used an exponentially decaying learning rate from 1e-3 to 1e-5 except for $\mathbf{m}_k$ in~\cref{eq:control_weights}, where we used a constant learning rate of 1e-2. We train for 20000 steps until it converges. Together with LRM generation of mesh sequence, this step takes around 30 minutes. In stage two, we used an exponentially decaying learning rate for all MLPs, including motion MLPs, appearance decoders, and pose encoders, from 1e-3 to 1e-5. We set constant learning rate of 1e-3 for Gaussian features on mesh vertices, Gaussian scales, and parent Gaussian opacities. We keep the constant learning rate of 1e-2 for $\mathbf{m}_k$. We train 400000 steps for the reconstruction to converge, which usually takes 60 minutes. In the initial 5000 steps, we train the model without Gaussian density control. After the initial 5000 steps, we use our adaptive density control technique mentioned in~\cref{sec:adaptive pop} to introduce new Gaussians and delete excessive Gaussians every 2000 steps until the final 5000 steps, when we stop the density control again.

\end{document}